# Online Tool Condition Monitoring Based on Parsimonious Ensemble+

Mahardhika Pratama, *Member, IEEE*, Eric Dimla, *Member, IEEE,* Tegoeh Tjahjowidodo, Witold Pedrycz, Fellow, IEEE, Edwin Lughofer

***Abstract*—Existing methodologies for tool condition monitoring still rely on batch approaches which cannot cope with a fast sampling rate of metal cutting process. Furthermore they require a retraining process to be completed from scratch when dealing with a new set of machining parameters. This paper presents an online tool condition monitoring approach based on Parsimonious Ensemble+ (pENsemble+). The unique feature of pENsemble+ lies in its highly flexible principle where both ensemble structure and base-classifier structure can automatically grow and shrink on the fly based on the characteristics of data streams. Moreover, the online feature selection scenario is integrated to actively sample relevant input attributes. The paper presents advancement of a newly developed ensemble learning algorithm, pENsemble, where online active learning scenario is incorporated to reduce operator's labelling effort. The ensemble merging scenario is proposed which allows reduction of ensemble complexity while retaining its diversity. Experimental studies utilising two real-world manufacturing data streams: metal turning and 3D-printing processes and comparisons with well-known algorithms were carried out. Furthermore, the efficacy of pENsemble+ was examined using benchmark concept drift data streams. It has been found that pENsemble+ incurs low structural complexity and results in a significant reduction of operator's labelling effort.**

***Index Terms*— Prognostic Health Management, Online Learning, Ensemble Classifier, Lifelong Learning, Nonstationary Environments, Concept Drifts**

## I. INTRODUCTION

TOOL condition monitoring (TCM) aims to feed real-time information of tool condition for the so-called maintenance on-demand framework where tool is replaced at the right time [1]. This paradigm brings cost saving to the industry because replacing sharp tools too early and too often incurs frequent shutdown of the machining process and leads to a dramatic increase of tool costs whereas worn tool potentially damages the surface finishing and dimensional integrity of work piece [2] and intensifies vibration level of the cutting process. When the tool is no longer at desired functionality or blunt, it entails high cutting force resulting in expensive energy cost [3].

The TCM usually involves two tasks, namely sensing and monitoring [5]. Sensing is a phase used to capture cutting signals from a set of sensors in the TCM. Sensing itself can be further classified into two types of modes, namely direct and indirect [6]. The second phase, namely monitoring, aims to perform predictive analytics from the measured signals. Existing monitoring approaches are categorized into three groups [4], namely first principle, data-driven and hybrid. The first-principle approach is impractical because of the fact that machining process is highly influenced by a number of dynamic factors: temperature, cutting fluids, chip formation, work piece and tool materials, etc. [54], [57]. Data-driven approach offers an alternative of the former one where predictive analytics is purely done using input/output data recorded by a number of sensors and a set of data acquisition units. This approach makes use of intelligent techniques which emulate dynamics of tool condition through a "learning" process of manufacturing data. Tool wear progression is measured by considering the plane-faced tool geometry approach with flank wear as the underlying variable of the tool life. The intelligent approaches feature generalization capability where it can be deployed to monitor tool condition in the real-time mode.

Recent progress in the TCM research has reported that the data-driven approach has gained increasing popularity in the community [12]-[15] because it can be deployed with a very-little capital expenditure [12]. It is done using exclusively sensory data and does not require complicated pre-setting requirements or assumption and/or simplification which is inherent in the first principle approach. The data-driven TCM method still requires more advanced data analytics because of at least two reasons: 1) existing approaches rely on a batch learning approach which is not fully compatible for online real-time processing; 2) existing approaches are constructed under a static structure predetermined before process runs. Such approaches are not self-adaptive, thereby being unable to adapt to variations of machining parameters.

As more and more industries have integrated the so-called

The first author acknowledges the support of NTU start-up grant and MOE Tier 1 Grant. The third author acknowledges and thanks to Mr. Lao Wenxin and Dr. Li Mingyang from Singapore Centre for 3D Printing, Nanyang Technological University, Singapore, for their support in the data collection. The fourth author acknowledges the support of the Natural Sciences and Engineering Research Council of Canada (NSERC) and Canada Research Chair (CRC) in Computational Intelligence. The fifth author acknowledges the support of the Austrian COMET-K2 programme of the Linz Center of Mechatronics (LCM), funded by the Austrian federal government and the federal state of Upper Austria. Mahardhika Pratama is with School of Computer Science and Engineering, Nanyang Technological University (e-mail: pratama@ieee.org). Eric Dimla is with the Mechanical Engineering Department, Faculty of Engineering, Boulder, University Teknologi Brunei, Brunei Darussalam (e-mail: dimla@utb.edu.bn). Tegoeh Tjahjowidodo is with Singapore Centre for 3D Printing, School of Mechanical and Aerospace Engineering, Nanyang Technological University, Singapore (email: ttegoeh@ntu.edu.sg). Witold Pedrycz is with the Department of Electrical and Computer Engineering, University of Alberta, Canada (email: wpedrycz@ualberta.ca). Edwin Lughofer is with the Department of Knowledge-Based and Mathematical Systems, Johanes Kepler University (email: Edwin.lughofer@jku.edu.at).

Internet-of-Things (IoT) in the current trend of automation and data exchange in the so called Industry 4.0 ($4^{th}$ industrial revolution), this calls for advancements of existing predictive maintenance to cope with online and dynamic characteristics of machining process [12]. The concept of Evolving Intelligent Systems (EIS) [42], [43] provides promising approach for online predictive maintenance because it features two important properties: online learning, dynamic and evolving structure. Vast majority of existing EISs are constructed under a single base-model where the evolving nature is generated from automatic partitioning of input and output space with fuzzy rule, neuron, etc. [44]-[50]. It is understood that the ensemble paradigm is capable of improving model's generalization because the classification decision is drawn from a collection of local experts. The ensemble method handles the bias-variance dilemma better than its single model counterpart provided that local experts exhibit good diversity. This advantage is normally achieved when incorporating weak local experts.

Despite being already mature as reported in the literature [51]-[53], most works in the ensemble learning scenario utilize non-evolving or even batched base classifier. This results in costly computational overhead and memory burdens. The use of evolving base-classifier helps the ensemble classifier to be more robust to deal with the local concept drift because it offers better exploration in the local region than those static classifiers. Few works in the literature [47]-[50] have incorporated evolving base classifiers under different ensemble configurations: bagging, boosting and stacking. These works are, however, crafted under a static ensemble structure which cannot adapt to the concept drift. Moreover, they suffer from the absence of drift detection scenario which identifies the presence of concept drift.

This paper presents a novel data-driven tool condition monitoring methodology benefiting from recent progress in the area of data stream analytics. An evolving ensemble classifier, namely Parsimonious Classifier+ (pENsemble+) is put forward. pENsemble+ handles aforementioned limitations because of the fact that it works fully in the single-pass fashion where data are directly discarded once learned without the requirement of secondary memory or archival storage. Furthermore, it adopts a fully evolving working scenario where both ensemble structure and base-classifier structure can be automatically generated and pruned from data streams. Moreover, the underlying innovation of pENsemble+ compared to its root, pENsemble [9] is implied by two facts: 1) pENsemble+ is equipped by the online active learning scenario which actively selects the training samples for model updates. This trait is vital for online tool condition monitoring because it relieves operator's annotation effort; 2) pENsemble+ introduces the notion of "ensemble merging scenario" which aims to maintain ensemble complexity in the low level while improving diversity of the ensemble classifier. This strategy offers an alternative of the significance-based pruning technique existing in the literature [10] which often compromises the model's diversity.

pENsemble+ is constructed with a generalized version of Dynamic Weighted Majority [11] which puts forward an open ensemble structure. Unlike the original DWM [11] and its extension in [41], pENsemble+ is equipped by an online active learning scenario which automatically selects training samples for model updates based on the Bayesian conflict measure which analyses conflict level in both feature space and target space. In realm of tool condition monitoring problem, the online active learning scenario resolves the major bottleneck of supervised learner which happens to be over-dependent on operator's feedback. pENsemble+'s structure is automatically generated using the drift detection method devised with the concept of Hoeffding bound [16].

pENsemble+ adopts the penalty and reward scenario where the base-classifier is punished when making misclassification, whereas a reward is granted provided it returns correct prediction. The reward scenario is an additional phase in respect to the original DWM meant to retain diversity of ensemble classifier. The diversity, however, must be interpreted with care in the data stream context which happens to be non-stationary because outdated or irrelevant classifier undermines final prediction. Complexity reduction scenario is incorporated with the ensemble merging scenario which focuses on the redundancy issue. The ensemble merging scenario offers plausible tradeoff between diversity and simplicity since it does not scan through poor classifiers rather focus on those redundant classifiers. Another unique feature of pENsemble+ is shown in its online feature selection scenario [17] which dynamically samples relevant input attributed during the training process. This mechanism assigns numeric weight (0 or 1) for every input attribute in every training observation and allows to arrive at different subsets of input attributes in the training process. pENsemble+ deploys an evolving fuzzy classifier, namely pClass as a local expert [18] which adopts an open structure. This provides additional flexibility in the base-classifier level. Furthermore, pENsemble+ will be implemented under the two variants of pClass, namely axis-parallel and multivariate. The difference between the two lies in the covariance matrix of the Gaussian function in the premise part.

The major contributions of this paper are outlined as follows:

1) The paper puts forward a new perspective for online tool condition monitoring approach based on a novel evolving ensemble classifier. The unique features of our approach are seen in its capabilities in handling the three bottlenecks of existing data-driven TCM, time and space complexity, concept drifts, and data uncertainty;

2) A novel ensemble classifier, namely pENsemble+, is proposed. This algorithm goes one step ahead when compared to its predecessor, pENsemble. The unique feature of pENsemble+ is shown in the online active learning scenario automatically sampling relevant samples for model updates and the ensemble merging scenario offering complexity reduction without compromising diversity of the ensemble classifier;

3) A real-world experiment in the metal-turning process was carried out where real-world manufacturing data were collected and preprocessed. Furthermore, another experiment in the 3D-printing process was undertaken in which the main goal was to perform the condition monitoring of 3D printer nozzles. The details of our numerical study in the 3D-printing process is placed in the supplemental document.

The efficacy of pENsemble+ is compared with a number of recently published algorithms: Learn++.NSE [19], Learn++.CDS [20], pENsemble [9] and pClass [18]. Additional numerical results are also served using popular concept drift problems in the literature. The advantage of pENsemble+ is evident in our experimental study where it attains significant improvement in time, space and sample complexity even compared to a single classifier without substantial compromise on accuracy. The paper is structured as follows: Section 2 encompasses learning policy of pENsemble+ and learning procedure of its base classifier, pClass; Section 3 outlines experimental procedure and machining setup; Section 4 elaborates on numerical results; and some concluding remarks are drawn in the last section of the paper.

## II. Learning Policy of pENsemble+

This section elaborates on fundamental working principle of pENsemble+ including ensemble learning mechanism and learning scenario of the base-classifier. An Overview of pENsemble+ learning mechanism is depicted in Fig. 1. pENsemble+ executes data streams on a chunk by chunk basis where each data chunk is fed to the online active learning scenario which is meant to shrink data chunk size and to relieve operator's labelling effort. The learning process continues with the online feature selection scenario which assigns numeric feature weights (0 or 1) for every input attributes. The performance of base classifier is evaluated based on its predictive performance reported on a new observation where misclassification triggers a penalty reducing its voting's weight while reward augments it. The complexity reduction scenario is implemented through the ensemble merging scenario. The drift detection scenario determines the learning stage which governs the stability and plasticity of the ensemble classifier.

### *A. Parsimonious Classifier (pClass)*

pENsemble+ deploys a newly developed evolving classifier, namely pClass as a base-classifier in order to attains greater flexibility in handling concept drift in individual local regions. This phenomenon is known as the local concept drift in the literature. It is evident that concept change applies to particular input regions only with different rates and severities.

- *Fuzzy Rule of pClass*: pClass is a class of first-order evolving fuzzy classifiers constructed with Takagi Sugeno Kang (TSK) fuzzy system where the rule consequent implements a first-order linear function while the rule premise is built upon the multivariate Gaussian function generating non-axis-parallel ellipsoidal clusters. Although the original pClass utilizes the multivariate Gaussian function with a non-diagonal inverse covariance matrix, the simplified version of pClass is also realized in pENsemble+ where a diagonal covariance matrix is used. This comparison aims to provide an overview of ensemble performance under two different base-classifiers.
- *Rule Growing Strategy of pClass*: pClass makes use of three rule growing modules, namely Datum Significance (DS), Data Quality (DQ), volume measure. The datum significance (DS) method is derived from the theory of statistical contribution presented in [21], [22]. This method aims to estimate the potential contribution of a rule or a data point during its lifespan. This method assumes that data samples are uniformly distributed and the statistical contribution is estimated using the Gaussian function as the kernel function. Because the DS method utilizes the uniform distribution assumption, it loses spatial information of data streams. This drawback is addressed in [23] by introducing the sliding-window-based approach where it calculates the accumulated firing strengths across data points in the sliding window. The size of sliding window is often problem-dependent. The DQ method is introduced to answer this bottleneck where it aims to extract density information of data samples. This strategy is inspired by the notion of recursive density estimation (RDE) in [24]. pClass extends this method for the multivariate Gaussian function and integrates the weighting function to cope with the outlier drawback [25]. Furthermore, the third rule growing module checks the volume of winning rule. It aims to limit the size of the fuzzy rules because the over-sized rules risks on the so-called cluster delamination problem. That is, one cluster may cover one or more distinct data distributions [26]. It must be noted that, as with pENsemble+, pClass is also implemented under the axis-parallel ellipsoidal rule in pENsemble+ to analyze the effect of two different base-classifiers. The same formulas as in the original pClass can be used except that the diagonal covariance matrix is deployed instead of the non-diagonal version.
- *Rule Pruning and Recall Strategy of pClass*: pClass is equipped by two rule pruning scenarios, namely Extended Rule Significance (ERS) and Potential+ (P+) methods. The ERS approach shares the same principle as the DQ method except the statistical contribution of a fuzzy rule is estimated instead of a data point. This component aims to scan through inconsequential rules which do not play significant role during its lifespan. Such rules can be pruned without compromising generalization performance. The P+ method, on the other hand, functions to capture outdated rules which are no longer relevant to represent current data distribution. This trait is made possible by inspecting the density evolution of fuzzy rules. The P+ method presents a modification of potential method in [27]. The potential method is used for the rule pruning scenario in [28] but the P+ method differs from this approach because it is based on the inverse multi-quadratic function in lieu of the Cauchy function. In addition, pClass incorporates the so-called rule recall scenario. This scenario allows previously pruned rules to be reactivated again in the future. This scenario refers to a case where old concept reappears again in the future. One can consider to introduce a new rule to overcome this situation but this strategy catastrophically erases past learning history. This situation is undesired because learning a local region must be restarted from scratch again.
- *Parameter Learning Scenario*: pClass utilizes the fuzzily weighted generalized recursive least square (FWGRLS) method which presents a weight decay term to retain small and bounded weight vector. This strategy is inspired by the concept of Generalized Recursive Least Square [29] which incorporates the weight decay term in the cost function of RLS [30]. The FWGRLS can be also seen as a variation of the fuzzily weighted recursive least square method [31] with addition of the weight decay term. The advantage of weight decay term is to safeguard the weight vector to keep its values small. It is worth noting that we adopt the simplified form of GRLS method where the second term is ignored. This leads to similar formulas of FWRLS method except the presence of weight decay term. This strategy is meant to improve the model's generalization and compactness of the rule base since a rule with a very small weight vector can be easily detected by the ERS method. There exist several types of weight decay term, say quadratic, quartic, multimodal, etc. The quadratic weight decay term is selected in the pClass since the weight vector proportionally decreases to their initial values. A flowchart of pClass learning procedure can be found in the supplemental document.

### *B. Ensemble Learning Scenario*

pENsemble+ is developed with a generalized version of DWM which adopts an open structure paradigm. This learning scenario clearly differs from original DWM at least in 4 facets:1) the voting weight is given a chance to increase and this strategy aims to retain diversity of ensemble classifier; 2) the drift detection strategy is deployed to introduce a new base-classifier whereas, in the original DWM, a new base-classifier is added when the global prediction returns misclassification; 3) the ensemble merging scenario based on the maximum information compression and the online feature selection scenario are absent from the original DWM. Algorithm 1 illustrates the fundamental working principle of pENsemble+.

pENsemble+ works on a chunk-by-chunk basis and if no base classifier exists, the first classifier is created using the first data chunk. The learning procedure starts with the online active learning scenario evaluating sample's contribution whether it deserves a learning process. The Bayesian conflict measure is deployed to measure conflicts in the input and output space. A sample is accepted for model updates provided that it satisfies dynamic sampling criteria. If a data sample meets the dynamic sampling criteria, the learning process continues with the labelling process followed by the online feature selection scenario. The online feature selection selects relevant input features by assigning crisp weights (0 or 1) and makes possible to arrive at different combinations of feature subsets in every training episode. The predictive performance of each base classifier is examined afterward where a classifier returning misclassification is penalized by decreasing its voting weight whereas a reward is given by increasing the voting weight when correct prediction is made The decreasing and reward factor, $p$, is selected at 0.5. The global prediction of ensemble classifier is inferred from a weighted sum of each class. The voting weight of each base classifier is normalized to allow proportional voting weights among each local experts. A class with the maximum weight is chosen as the predicted class. This procedure is followed by the ensemble merging procedure which is meant to capture redundant classifiers. Two classifiers with high mutual information are coalesced. The last phase of the training procedure is the drift detection scenario categorizing dynamic of data streams into three conditions, stable, warning and drift. When a drift is signaled, a new base-classifier is introduced. No action is performed during the warning condition since this phase depicts a transition period before a drift is confirmed. Such situation usually occurs in the presence of gradual drift. The winning classifier is updated using the newest data chunk during the stable phase to keep up-to-date with the most recent concept and to prevent over-fitting. The winning classifier is selected from that having the lowest predictive error – MSE. The learning components of pENsemble+ and relevance of pENsemble+ for TCM are detailed as follows:

• *Online Active Learning Strategy*: online active learning scenario is urgently required in the complex manufacturing process because of the cost in obtaining the true class label. This usually requires a complete shutdown in the machining process since the flank wear has to be evaluated through visual inspection or at least some delay is expected to receive the true class label. pENsemble+ features an online active learning scenario based on the extended conflict ignorance (ECI) paradigm [32] which evaluates conflict in both feature and target domains. This strategy was derived from the conflict and ignorance method for the conventional TSK fuzzy classifier [33] where the underlying difference lies in the use of a dynamic sampling paradigm [34] and Bayesian posterior probability estimation in both input and output space. None of these works, however, investigate the ECI method within the context of ensemble classifier. As a matter of fact, the ensemble learning scenario requires an innovation for sample evaluation strategy since it consists of a collection of local experts evolved from different data space. One must start from the fact that a data sample may incur different conflict degrees in different local experts. Although the sample evaluation strategy should take place at the local level, a centralistic sample evaluation strategy where all base classifiers are put together under one roof to produce the predicted class label is formed. This strategy is chosen to suit the online feature selection module of pENsemble+ which also adopts the centralistic feature selection scenario. Moreover, it is found that this does not make substantial difference since the maximum operator has to be ultimately committed when performing local sample evaluation to analyze the confidence of ensemble classifier. The advantage of centralistic sample selection is evident in the stability aspect because it does not depend on the performance of local learner.

The Bayesian conflict measure in both input and target space is utilized to evaluate conflict level for each local expert. The Bayesian approach is preferred over a standard distance or firing strength measure because it encompasses the prior probability and the joint –category and class probability. A sample is conflicting not only because it is out of scope of a current fuzzy rule but also if it occupies "unclean" region shared different class samples. Moreover, the prior probability is required to take into account the cluster's population since a highly populated cluster tends to be "frozen". That is, it is no longer responsive to accept new training stimuli due to the characteristic of rule premise update affected by the cluster's support. The conflict in the output space, on the other hand, is measured from the classifier's truncated output. The classifier's output here is taken from the preference degree [32] determined with respect to the two most dominant classes since it intuitively informs about the degree of closeness to the decision boundary. A sample is conflicting if it falls near two-decision boundary and results in the preference degree with the value around 0.5.

The Bayesian conflict measure formed in the input space is expressed as follows:

$$P(y_o|X)=\frac{\sum_{i=1}^{R}P(y_j|R_i)P(X|R_i)P(R_i)}{\sum_{j=1}^{o}\sum_{i=1}^{R}P(y_j|R_i)P(X|R_i)P(R_i)} \qquad (1)$$

where $P(y_o|X), P(X|\overline{R}_i), P(X|\underline{R}_i), P(R_i)$ respectively stand for the joint-class and category probability, the upper likelihood function, the lower likelihood function and the prior probability. The joint-class and category probability and the prior probability are defined as follows:

$$P(R_i)=\frac{N_i}{\sum_{i=1}^{R}N_i},\ P(y_o|R_i)=\frac{N_{i,j}}{\sum_{j=1}^{o}N_{i,j}} \qquad (2)$$

Algorithm 1. Pseudocode of pENsemble+

**Algorithm 1:** Parsimonious Ensemble+ (pENsemble+)

Given a data chunk $D=(X^{P\times n},T^{P\times O})$ where $P,n,O$ are chunk size, the number of input dimension, the number of output dimension; set the adjustment factor $p_i$, the pruning threshold $\theta$; $\hat{C},\lambda,\sigma\in\Re^{1\times O}$ are global and local predictions, sum of weighted predictions for each class

a data chunk $D\in\Re^{\mathrm{P}\times(n+O)}$ is received
**For** $t=1,\ldots,\mathrm{P}$ // loops over all examples in the data chunk
**IF** the ensemble network is empty
$M=1$ // create the first local expert, $\beta_i=1$ // initialize the weight of a local expert
**End**
$\sigma=0$
Execute the online active learning (1)-(4)
**IF** $P(y_0|X)^{input}>\theta$ **OR** $P(y_0|X)^{output}>\theta$
Discard the data sample
**ELSE**
Accept and label the data sample for model update
Execute the feature selection mechanism (7)-(9)
**For** $i=1,\ldots,M$ // loop over local experts
$\lambda=\max_{j=1,\ldots,O}(y_{i,j})$ // elicits the local prediction
**IF** ( $\lambda\neq C^t$ )
$y_i=\beta_i y_i$ // decreases the weight of a local expert when it predicts incorrectly
$\beta_i=\beta_i p$
**Else**
$\beta_i=\min(\beta_i(2-p),1)$
**End**
$\sigma_\lambda=\sigma_\lambda+y_i$
**End**
$\hat{C}=\max_{\lambda=1,\ldots,O}(\sigma_\lambda)$ // Produces the global prediction, $\beta_i=\frac{\beta_i}{\sum_{i=1}^{M}\beta_i}$ // normalizes the weight
**For** $i=1,\ldots,M$
Undertake the ensemble merging procedure based on the maximum correlation index
**IF** (12)
Discard *i-th* local expert
**End**
**End**
Undertakes the drift detection method (13),(14)
**IF** Drift
Introduces a new base classifier
**ElseIF** Warning
Do nothing and prepare for possible drift in the next observation
**ElseIF** Stable
Train the winning Classifier
**End**
**End**

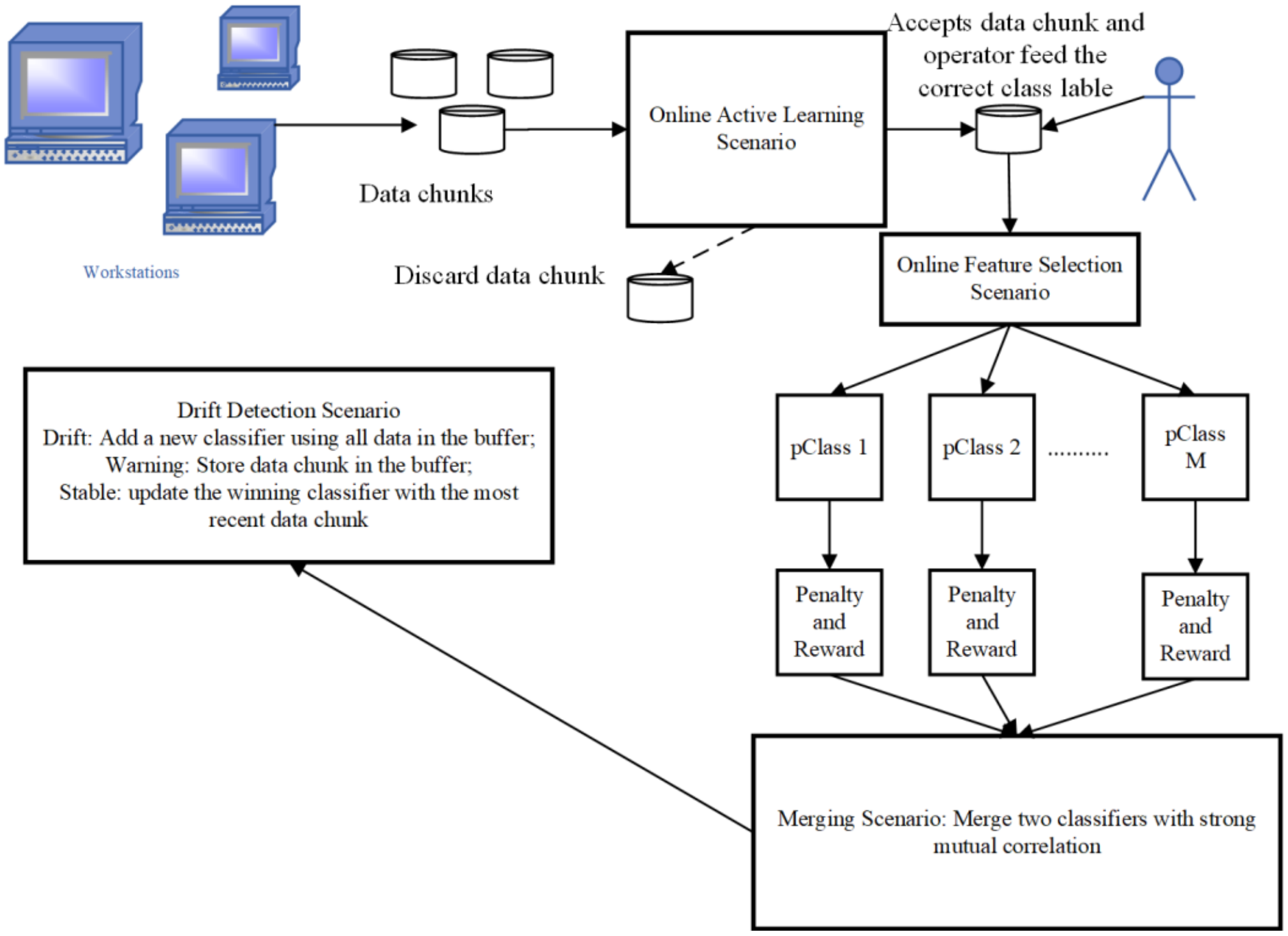

Fig. 1 Learning Policy of pENsemble+

where $N_i, N_{i,j}$ stand for the support of *i-th* rule and the support of the *j-th* class of the *i-th* cluster. (2), (3) can be softened by adding the *log* operation. This approach is useful in the category choice phase because it provides higher likelihood for a newly created cluster to be selected as the winning rule. The joint-class and category probability is estimated by the number of *j-th* class of the *i-th* rule which signifies the purity degree of a fuzzy rule [18]. The class overlapping condition is most likely to be found in the case of unpurified cluster. The likelihood function can be defined in the similar way as their crisp version using the Mahalanobis distance as follows:

$$P(X|\tilde{R}_i) = \frac{1}{(2\pi V_i)^{1/2}} \exp(-(X-C_i)\Sigma_i^{-1}(X-C_i)^T) \quad (3)$$

where $C_i, \Sigma_i^{-1}$ denote the Center and inverse covariance matrix of the *i-th* rule while $V_i$ stands for the volume of *i-th* rule. The advantage of Bayesian approach is also clear when fuzzy rules occupy in almost similar proximity to a data sample because of its prior probability. The volume of fuzzy rules can be obtained with ease with the determinant operator. If a precise estimation is required, it can be calculated as shown in [18] where the Gamma function and the eigenvalue are utilised. Note that the determinant may return negative volumes as per its definition as the signed volume.

The conflict in the output space is evaluated from the classifier's truncated output to guarantee that it lies in in the interval [0,1] as follows:

$$P(y_o|X) = \min(\max(conf,0),1), conf = \frac{y_1}{y_1+y_2} \quad (4)$$

where $y_1, y_2$ stand for the most and second most dominant classes which can be obtained from the highest and second highest outputs of pClass. It is worth mentioning that pClass characterizes the regression-based classifier constructed under the MIMO architecture which scatters rule consequent for each output. It is evident that this formula portrays the classifier's confusion perfectly because a significant conflict is indicated when a classifier does not produce a conclusive prediction – two outputs have about the same values. This situation may occur when a data sample is geometrically close to the decision boundary separating the two classes. Suppose that $P(y_0|X)^{input}$, $P(y_0|X)^{output}$ are the estimate of posterior probability in input space and output space, the condition of sample acceptance is formalized as follows:

$$(P(y_0|X)^{input} \le \theta \text{ or } P(y_0|X)^{output} \le \theta) \quad (5)$$

where $\theta$ is the conflict threshold. The higher the value of this parameter the higher the number of training samples are accepted for model updates, whereas the lower the value of this parameter the fewer the number of training samples are discarded for model updates. A sample is supposed to be a good candidate for model updates, if it results in significant conflict for all base classifiers. This strategy aims to enhance the diversity of the ensemble classifier by preventing redundant samples to be learned. Furthermore, a budget [34] controlling the maximum labelling cost can be inserted in (5)-(7). This approach is useful when the true class label is too expensive to be obtained such as in the bioinformatics applications. The online active learning can also function to relieve the class imbalance issue. First, the imbalance factor is estimated to find the minority and majority classes. The online active learning is set loose for minority class samples up to a point where a balanced proportion of target classes has been achieved. That is, minority class samples are always sampled to attain equal class distribution. Since the true class label is unknown in realm of online active learning scenario, it is estimated with the posterior probability in the input and output space (1), (4).

• *Online Feature Selection Scenario*: the online feature selection scenario is based on the OFS method in [17]. The OFS method is generalized here to be well suited to the ensemble working scenario since the original version only covers its implementation to the single linear regression. The unique feature of this approach lies in the fact that it makes possible to arrive at different subsets of input attributes by assigning binary weights (0 or 1). In other words, it removes the risk of discontinuity because it provides likelihood for every input feature to be selected in every observation. The OFS method cannot be directly implemented in the ensemble learning scenario because sensitivity of input attributes should be analyzed with respect to all base classifiers. This issue leads to carry out this scenario in a centralistic manner. That is, all fuzzy rules of base-classifiers are put together to perform the OFS procedure. Note that this mechanism is made possible by the fact that pClass adopts a local learning scenario where every rule is loosely coupled and has its own output covariance matrix. Let's recall the fuzzy rule of pClass comes as follows:

$$R_i : IF\ X_n \text{ is } N(X;C_i,\Sigma_i^{-1})\ THEN\ y_i^o = x_e W_i^o \tag{6}$$

where $C_i \in \Re^u$ is the numeric center of the multivariate Gaussian function and $u$ is the number of input variables. $\Sigma_i^{-1} \in \Re^{u\times u}$ is the inverse covariance matrix, $x_e \in \Re^{(u+1)}$ is the extended input vector, $W_i^o \in \Re^{(u+1)}$ is the output weight vector.

The OFS procedure starts by examining the prediction of a single pClass created by all rules of ensemble classifiers – all fuzzy rules are put together to construct a single classifier. The OFS method only takes place when a model returns misclassification $C \neq \hat{C}$ to save computational cost because the OFS method is meant to recover predictive quality of the model by getting rid of the influence of poor features. The output weight vector $W_i^o$ is adjusted using the stochastic gradient descent approach as follows:

$$W_i = W_i - \alpha\chi W_i - \alpha\chi\frac{\partial E}{\partial W_i} \tag{7}$$

where $\alpha, \chi$ are learning rate and regularization factor, respectively. The gradient term $\frac{\partial E}{\partial W}$ can be derived with ease by applying the standard MSE as the cost function. The stochastic gradient descent is utilized to adjust the output weight rather than the FWGRLS method as pClass, since the OFS method is undertaken in the centralistic manner having different optimization objective from that of the base-classifier level. Moreover, the stochastic gradient descent is much easier to be executed than that of the FWGRLS method because no output covariance matrix has to be assigned when performing the OFS scenario.

To guarantee a bounded norm, the output weight vector is projected to the $L_2$ ball as follows:

$$W_i = \min\left(1, \frac{1/\sqrt{\chi}}{\|\tilde{W}_i\|_2}\right)W_i \tag{8}$$

This strategy is also required to examine whether values of the output weight vector is concentrated within the $L_2$ ball and thus pruning small values, being remote from the $L_2$ ball center, does not compromise the model's generalization. The contribution of input attribute is informed from its dominance in the output weight vector. In realm of the TSK fuzzy system, the rule consequent or the output weight vector steers the direction or the tendency of a rule in the output space. In addition, the output weight vector is more stable than the gradient information (changing in each observation) during the sensitivity analysis to inform direction of predictive tasks. The contribution of input attribute is expressed in the form:

$$I_j = \frac{\sum_{i=1}^{R} W_{j,i}}{\sum_{j=1}^{u}\sum_{i=1}^{R} W_{j,i}} \tag{9}$$

where $I_j$ is the sensitivity of $j$-th input feature and $R$ is a total number of fuzzy rules across all base-classifiers. Note that the data standardization must be performed in this context because different input ranges obscure the true contribution of input attributes. Suppose that $B$ is the desired input dimensionality and $B$ is smaller than the original input dimension $u$, the input attributes with the $B$ largest input contributions $I_j$ are picked up in every observation and the remainder of input attributes are ruled out from the training process by assigning 0 weights. Input attributes are not permanently forgotten and are reactivated in the future whenever they are called for the current data distribution – cyclic drift. The OFS method also covers the case of partial input information required when the cost of feature extraction is too costly. Because the partial input information is similar to that of full input version, it is not recounted in this paper. The sensitivity measure (9) is also used in eTS+ [55]. Nonetheless, eTS+ adopts the hard input pruning mechanism where superfluous features are permanently discarded without any opportunity to be picked up again.

• *Ensemble Pruning Scenario*: The main bottleneck of ensemble classifier for data stream application is found in the issue of computational and space complexity because it incurs considerable complexity if it consists of a large collection of base classifiers. Nonetheless, the ensemble pruning scenario is often counterproductive for classifier's accuracy since it limits the diversity of the ensemble classifier [35] – the underlying strength of the ensemble classifier. The ensemble pruning scenario discards superfluous classifiers – either poor classifier or outdated classifier. Although such classifiers play little during their lifespan, they remain important to generate diverse output space. In realm of dynamic and evolving learning environments, significance of base classifiers usually changes rapidly in accordance to the context. When using the ensemble pruning scenario, it is necessary to integrate the recall capability because already pruned classifiers may turn out to be useful again to cover future data distribution. The most plausible approach to complexity reduction of ensemble classifier is by putting forward the ensemble merging scenario. That is, it analyses mutual information of base classifiers and base classifiers featuring strong mutual information are merged to be a single classifier. The mutual information is quantified by comparing the classification output of two classifiers. The mutual information is relatively more stable than the significance-based criterion because it measures correlation between two base classifiers.

An Analysis of mutual information can be performed using any correlation measure provided that they satisfy the online learning requirements. Nonlinear correlation measure is often more accurate than linear correlation measure but it is not scalable for online real-time processing and requires simplified assumptions such as training samples follow normal distribution [36]. pENsemble+ utilizes the maximal compression index (MCI) to measure correlation of two base classifiers. This approach is more robust than conventional Pearson correlation index since it is insensitive to rotation and translation [37]. It calculates the amount of information loss when ignoring one of the base classifier and if no significant difference exists or MCI is small, the information of ignored base classifier is already covered by its pair. The MCI is expressed as follows:

$$\xi(y_1,y_2)=\frac{1}{2}(\mathrm{var}(y_1)+\mathrm{var}(y_2)) \\ -\sqrt{(\mathrm{var}(y_1)+\mathrm{var}(y_2))^2-4\,\mathrm{var}(y_1)\,\mathrm{var}(y_2)(1-\rho(y_1,y_2)^2)} \tag{10}$$

$$\rho(y_1,y_2)=\frac{\mathrm{cov}(y_1,y_2)}{\sqrt{\mathrm{var}(y_1)\,\mathrm{var}(y_2)}} \tag{11}$$

where $\mathrm{cov}(y_1,y_2),\mathrm{var}(y_1),\mathrm{var}(y_2)$, $\rho(y_1,y_2)$ denote the covariance of classifier's outputs $y_1,y_2$, variance of classifier's output $y_1$, variance of classifier's output $y_2$, and the Pearson correlation index, respectively. The variance and covariance can be calculated recursively with ease. It is worth mentioning that the MCI satisfies the following properties: 1) $0\le\xi(y_1,y_2)\le 0.5(\mathrm{var}(y_1)+\mathrm{var}(y_2))$; 2) the maximum correlation is attained when $\xi(y_1,y_2)=0$; 3) $\xi(y_1,y_2)=\xi(y_2,y_1)$; 4) it is insensitive against the translation because mean expression is nowhere in (6); 5) it is insensitive to rotation because a perpendicular distance of a point to a line is not dependent on rotation. The ensemble merging condition is set as follows:

$$\xi(y_1,y_2)<\delta \tag{12}$$

where $\delta$ is a merging threshold. The lower the value of this threshold implies less merging process to be performed in the training process, whereas the higher the value of this threshold induces more aggressive merging process is committed during the training process. Once a merging decision is taken, one of the two classifiers is discarded. The classifier with a lower accuracy is selected for the pruning process while another one is retained.

- *Drift Detection Scenario*: the dynamic of pENsemble+ is controlled by a drift detection scenario, which aims to discover abnormal patterns leading to possible change of data stream dynamics. The drift detection is based on the Hoeffding's inequalities [16] which classifies dynamics of data streams into three categories, namely normal, warning and drift. The normal phase means no variation in data streams is found, while change still needs for further investigation in the warning phase. The drift phase means that change is certain in data streams. The advantage of this method lies in assumption-free for the probability density function. It assumes data streams as independent and bounded random variables.

The drift detection scenario is carried out by inspecting statistics of data streams – moving average - without any weight. Although its weighted moving average variant does also exist in [16], the standard moving average is deployed here since it is more sensitive to abrupt change than the weighted version and also is easy-to-use because it does not call for specific tuning scenario for weight adjustment. The statistics of data streams is computed as $\hat{X}_t=\sum_{t=1}^{\mathrm{P}}\Upsilon_t X_t,\Upsilon_t=1/\mathrm{P},\hat{X}=\bar{X}$. Note that its recursive version can be derived with ease. The drift detection strategy adopts similar concept as the statistical process control except the assumption of normal distribution is removed. The standard deviation in the confidence interval $\sigma$ is replaced with the significance level $\alpha$. Two significance levels are implemented to determine the conflict level in data streams. They correspond to two levels of the drift detection scenario, namely the warning level ($\alpha_W$) and to the drift level ($\alpha_D$).

This strategy partitions a data chunk into three groups, namely $X=[x_1,x_2,...,x_{\mathrm{P}}]\in\Re^{\mathrm{P}\times u}$, $Y=[x_{cut+1},x_{cut+2},...,x_{\mathrm{P}}]\in\Re^{\mathrm{P}-cut+1\times u}$, $Z=[x_1,x_2,...,x_{cut}]\in\Re^{cut\times u}$ where $\hat{X},\hat{Y},\hat{Z}$ are statistics computed from these three data groups. Each group is assigned with a Hoeffding's error bound $\varepsilon_X,\varepsilon_Y,\varepsilon_Z$ to set proper conflict levels which signify the status of data streams. The error bounds are allocated as follows:

$$\varepsilon_\alpha=(b-a)\sqrt{\frac{(\mathrm{P}-cut)}{2cut(\mathrm{P}-cut)}\ln(\frac{1}{\alpha})} \tag{13}$$

where $[a,b]$ are the minimum and maximum values of input attributes and $\alpha$ is the significance level. Note that the significance level has a clear statistical interpretation which corresponds to the confidence level of the Hoeffding's bounds $1-\alpha$.

The drift detection starts by finding the cutting point which pinpoints a switch point of two data distributions and in turn partitions the data chunk into the three groups. The switch point does not signal directly drift condition to prevent the outlier's effects rather in-depth investigations must be performed to ascertain the status of data distribution – whether a drift really presents. It is in line with the fact of gradual drift where three exists a transition period that depicts a mix between two distributions. A data point is said to be a cut point given that the following condition is met.

$$\widehat{Z}_t + \varepsilon_{\widehat{Z}_t} \geq \widehat{X}_t + \varepsilon_{\widehat{X}_t} \quad (14)$$

A point indicating the mean increase is sought as the cutting point. The cutting point partitions a data chunk into two groups compared to confirm the existence of a drift.

The next step is to determine the status of data streams by formulating a hypothesis test. If null hypothesis is rejected with size $\alpha_D$, a drift's status is returned, whereas a warning status is indicated, if null hypothesis is rejected with size $\alpha_D, \alpha_W$. In other words, $\alpha_D, \alpha_W$ correspond to the confidence level of Hoeffding's inequality for different stages: $1-\alpha_W$ (warning), $1-\alpha_D$ (drift). The lower the value of $\alpha_D, \alpha_W$ the lower the confidence level of Hoeffding's bound is – more examples are considered to be a drift. This implies to more base-classifiers to be added during the training process and vice versa. The null hypothesis is formed as $H_0: E(\widehat{X}) \leq E(\widehat{Y})$, while its alternative is defined as $H_1: E(\widehat{X}) > E(\widehat{Y})$. Since the weight is excluded here, this hypothesis is analyzed as $|\widehat{X} - \widehat{Y}| \geq \varepsilon_\alpha$ where $\varepsilon_\alpha$ is found from (10) by applying a specific significance levels $\alpha_D, \alpha_W$ which correspond to either drift or warning. This hypothesis inspects the dynamic of data streams after a switching point. This is meant to substantiate the presence of drift in data streams. If the null hypothesis happens to be maintained, the stable condition is signaled.

- *pENsemble+ for Online TCM Problems:* there exist at least three challenges in the TCM applications as follows:

1) Continual Machining Process: the online TCM problem cannot be handled in the offline manner meaning that a complete dataset has to be fully iterated over a number of epochs and a retraining phase from scratch has to be imposed when there is a changing pattern of data stream.
2) Non-stationary Working Environments: different machining parameters, namely cutting speed, feed rate, etc., and cutter profiles have to be applied to suit different product designs and shapes. In addition, aging components, changing tool condition, surface integrity, and external disturbances also affect machining conditions and have to be carefully taken into account to arrive at accurate monitoring of tool and surface quality. In other words, TCM is inherent to the concept drift which calls for not only online parameter tuning of machine learning but also an open structure property which allows to follow variations of system dynamics.
3) Limited Operator Intervention: a fully automated machining process is highly desired in the manufacturing process because it prevents possibility of operator's fatigue and errors. An unsupervised learning scenario is suitable for TCM application but an accurate prediction is difficult to be attained because of the absence of ground truth notably if it deals with a diagnostic phase where a fault has to be associated with classes.

pENsemble+ is relevant for TCM applications because it features an online working principle and a fully open structure in both local model and ensemble levels. The drift detection mechanism is devised to expand the ensemble structure once a drift is identified while an ensemble merging mechanism is put forward for complexity reduction purposes. Furthermore, the ensemble structure is composed of pClass – an evolving classifier supposed to handle local drift due to its dynamic structure attribute. The online active learning scenario is a concrete answer to the demand of limited operator intervention and ground truth because it compresses the original problem space. A compressed problem space having fewer data samples than the original ones significantly reduces operator's labelling cost. In other words, an incoming sample is vetted by the online active learning strategy where only accepted samples are passed to operators for labelling efforts. The online feature selection scenario bypasses conventional feature selection scenario in the conventional TCM executed in the offline and batched mode. This principle supports more adaptive TCM approach because feature contribution changes overtime.

## III. Experimental Set-up & Data collection

The set-up consisted of a Lang Swing J6 centre-lathe, onto which a Kistler tool-post dynamometer platform (type 9263A) was mounted to measure three mutually perpendicular components of cutting force. A Kistler tri-axial accelerometer (type 8730A) was used to measure three mutually perpendicular components of vibration from the underside of the tool holder. A Kistler charge amplifier (type 5006) and a Kistler power supply/coupler (type 5134) were used to amplify and decouple the cutting force and acceleration signals. Dry cutting was carried out on EN24T BS 970 817M40 alloy steel work-piece of Brinell hardness 255 of composition: 0.4% C, 0.28% Si, 0.27% Mo, 1.18% Cr, 0.5 % Mn and 1.4% Ni. The tool holder was Sandvik SSBCR 2020 K12 and the throwaway inserts were Sandvik Coromant of type SCMT 12 04 08 UM and material P25 4025 and P15 4015. Three cutting speeds (m/min) and feed-rates (mm/rev) of 275, 300, 350 and 0.1, 0.2 & 0.3 respectively were used at a constant depth of cut of 2 mm. Six signals were recorded using an Amplicon PC-30 data acquisition card mounted in a personal computer, at 4096 (N) samples per channel, a sampling rate of 30 kHz, and recorded on computer for analysis. It is worth mentioning that machining parameters here are set to suit the desired surface roughness and regularity of work piece. Different machining parameters induce the so-called concept drift. The incremental drift occurs in this case because the values of cutting speeds and feed rates incrementally increases. Moreover, the force and vibration signals characterizes noisy characteristic which causes the problem of data uncertainty. The feature extraction step is crucial to generate accurate features which underpins accurate identification of tool wear. The characteristic of fuzzy system also helps in dealing with noisy samples since its remote position always results in low membership degrees.

In most metal turning processes, flank and crater wear are usually the more prevalent forms of tool wear when cutting with plane-faced geometry inserts, and their occurrence are unavoidable. A judicious choice of cutting conditions can remedy other forms of wear such as frittering, notch and nose. In this investigation, coated (P15) and un-coated (P25) carbide inserts with chip breaker geometry to reduce chip/tool contact were utilized in order to minimize crater wear. The process was interrupted occasionally to record the flank/nose wear lengths measured with the aid of a Tool Maker's Microscope. The ISO 3685 wear criterion was used as a guide but not strictly applied. The decision for percentage wear on the cutting tool used as the classification benchmark as either worn or sharp was rather subjective. This was because each test cut began with a fresh tool insert and cutting continued until either the tool failed or flank/ nose wear had accumulated excessively. The following guideline was employed in determining the tool class:

i. fresh (sharp), flank wear $< 0.1$ mm;

ii. nominally sharp, 0.1 mm $\leq$ flanks wear $< 0.15$ mm;

iii. partly worn, , 0.12 mm $\leq$ flanks wear $< 0.15$;

iv. worn, 0.15 mm $\leq$ flanks wear $< 0.17$mm;

v. severely worn, flanks wear $\geq 0.17$ mm.

The flank wear values are determined from its effect to the surface quality and other equipment. It is well-known that blunt tool imposes higher cutting force which leads to costly scrap and even damages made to the machine.

The cutting conditions were incorporated to the input vector sets to assure that the underlying process parameters would be less sensitive to changes in the cutting conditions (cutting speed, feed rate and depth of cut). In total, 12 input time-domain and frequency-domain attributes are collected to identify the five conditions of the tool and comprises cutting speed, feed rate, depth of cut, static force in three axes X, Y, Z, dynamic force in three axes X, Y, Z and vibration in three cutting axes X, Y, Z. The TCM problem is formulated as a binary classification problem where fresh, nominally sharp and partly worn are grouped as a healthy class (class 0) while worn and severely worn are categorized as a worn class (class 1).

## IV. DATA PROCESSING AND SENSOR FUSION

Any typical sensor is used to measure one desired parameter and any other parameters influencing the measurement are considered to be interfering with the measurement. For example, a dynamometer is often used in measuring cutting forces (or thereof moments) which are then correlated to the process of interest. Because of the complex and non-linear nature of the cutting process, sensor co-operation is not only desired but also necessary. Essentially, sensor fusion relies on the fact that the fusion of different source signals of probable mediocre quality, yields better results than when only one such a signal is used [1] [2]. A comprehensive review of the synergistic integration of multi-sensor information can be found in [3] with typical application scenarios in TCM found in [4-6] [7]. Sensor fusion traditionally has been performed through application of statistical methods such as PCS or regression analyses [8] or a set of heuristic rules, but machine learning approach has recently gained popularity for TCM.

The instrumentation set-up did not allow simultaneous recording of the static and dynamic cutting forces. The parameters were mathematically extracted from the recorded cutting force signals. The dynamic forces were found by calculating the oscillatory part of the sampled force signals, meanwhile the arithmetic mean of the sampled cutting force signals were taken to represent the static cutting force components. The obtained dynamic data (dynamic force and vibration) were passed through a forward FFT, and the DC component of the dynamic force eliminated. The sum total power contained in the FFT spectrum (N/2) was taken for each cut and the values from this calculation together with the static forces formed the input data samples. In total, there were 157 data samples of which less than 10% could be said to represent the worn tool – partly worn, worn, severely word, while the rest were for a nominally sharp tool – fresh and nominally sharp. The obtained data were normalized such that the distribution for each signal was positioned within the 0.1-0.9 range, had zero mean with an equal variance distribution.

Table 1. Numerical Results

| Algorithms | Case | CR | FR | BC | NP | IA | TS | RT |
|---|---|---|---|---|---|---|---|---|
| pENsemble+ (multivariate) | Binary | **0.72±0.4** | **1.3±0.6** | **1.3±0.6** | **16±6.9** | **2** | 8.3±1.5 | **0.04±0.02** |
| pENsemble+ (axis-parallel) | | **0.72±0.4** | 2.67±1.15 | **1.3±0.6** | 26.6±11.5 | **2** | **8±0** | **0.04±0.02** |
| pENsemble | | **0.72±0.4** | **1.3±0.6** | **1.3±0.6** | **16±6.9** | **2** | 10 | 0.06±0.04 |
| pENsemble+ (multivariate) | Multiclass | **0.71±0.2** | **1** | **1** | 18 | **2** | 8.4±0.9 | 0.04±0.01 |
| pENsemble+ (axis-parallel) | | 0.52±0.16 | 1 | 1 | **16** | **2** | **8.4±0** | **0.03±0.01** |
| pENsemble | | 0.61±0.13 | 1 | 1 | 18 | **2** | 10 | 0.08±0.08 |

## V. PREDICTION PHASE

The TCM study case is presented as the binary classification problem where the underlying goal is to identify two tool conditions: healthy and worn derived from the five tool conditions as presented in Section 5. In addition, the multi-class TCM problem is used to numerically validate the performance of pENsemble+. Unlike the binary TCM problem in which the tool condition is identified from the flank wear, the multi-class problem forms a fault diagnosis problem due to three causes: flank wear, nose wear and fractured nose. It results in six tool conditions: nominally sharp, high flank wear, high nose wear, chipped/fractured nose, high flank and high nose wear, high flank and fractured/chipped nose. Four-class classification problem is formed based on the following guideline as follows:

- Flank wear mark value $\leq 0.15$mm, tool insert nominally sharp
- Flank wear mark value $> 0.15$mm, tool insert worn (high flank)
- Nose wear length $\leq 0.2$mm, nominally sharp
- Nose wear length $> 0.2$mm, tool worn (nose fractured / chipped)

The experimental set up and data collection procedure of the multi-class problem is detailed in the supplemental document. The multi-class dataset relies on the 12 features where the cutting speed varies (m/min) between 876 and 786 while 0.25, 0.5, and 0.75 feed rates (mm/rev) are applied. The depth of cut is set constant at 5.71 mm.

To simulate continual learning environments, our simulation follows the periodic hold out process also known as the train-then-test approach. That is, all data points are equally partitioned into a number of data chunks as the number of time stamps. Each chunk consists of two data group, namely training and testing samples where each model is updated first using the training samples and tested afterward using the testing sample. This procedure is repeated until data chunks have been all streamed. The numerical results are calculated from the average numerical results across all data chunks. In other words, the performance in handling each chunk is recorded and averaged after completing the hold-out procedure. Table 3 summarizes details of our numerical study including characteristics of datasets and the experimental procedure. Because the class imbalance is frequently encountered in the TCM domain [55], the imbalance factor is indicated in Table 3 [56].

pENsemble+ was benchmarked against pENsemble which happens to be predecessor of our proposed algorithm to exhibit to what extent the proposed methodologies in this paper are capable of improving numerical results of pENsemble. pENsemble characterizes a fully evolving fuzzy classifier using pClass [18] as a base classifier. It, however, suffers from the absence of online active learning scenario and still utilizes the generalization-based ensemble pruning scenarios. pENsemble+ is implemented with two types of pClass, axis-parallel and multivariate, to perceive the effect of base-classifier to the overall learning performance. The difference between the two is seen in the rule premise from which two different ellipsoidal clusters are generated automatically. Our simulation is carried out in MATLAB under a laptop with Intel Core i7-6500U CPU and 16 GB of RAM. The MATLAB codes of pENsemble+ are provided in[1)]. Two benchmarked algorithms are evaluated in 5 criteria: classification rate (CR), Fuzzy Rule (FR), base classifier (BC), network parameters (NP), input attribute (IA), training samples (TS), and runtime (RT). Numerical results are tabulated in Table 1.

It is seen from Table 1 that significant difference can be seen in terms of predictive accuracy where pENsemble+ classifies data samples with higher accuracy than pENsemble. It is also noted that the active learning scenario contributes to lowering the number of data samples to be labelled. Another salient observation is present in the use of multivariate Gaussian rule which delivers higher classification rate than the axis-parallel rule. Since the multivariate Gaussian rule has non-zero off diagonal elements, it incurs higher number of network parameters than the axis-parallel rule. Additional numerical study with TCM problem of the metal turning problem can be found in the supplemental document. In addition, the supplemental document serves TCM problem of 3D-printing process from our real-world project.

## VI. Numerical Study with Benchmark Problems

This section elaborates on numerical study using three popular concept drift problems, namely hyperplane, SEA and SUSY. pENsemble+ is compared with the same set of algorithms and is evaluated with the same performance metrics. As with the TCM problem, the experimental procedure follows the periodic hold-out process. This procedure simulates real data stream environments where past samples are discarded once seen. Table 2 shows numerical results of the consolidated algorithms.

A) *Hyperplane Problem*: this problem is obtained from the data stream generator of massive online analysis (MOA) [38]. It features a binary classification problem where the main task is to classify a data point into two classes with respect to the random hyperplane in the $d$-dimensional feature space. A class is classified as the positive class if $\sum_{i=1}^{d} w_i x_i > w_0$ , whereas the negative class is resulted from $\sum_{i=1}^{d} w_i x_i < w_0$ . The unique property of this problem lies in the gradual concept drift where at first data are drawn from one distribution with probability one. This concept gradually shifts to another data distribution up to the point where the second concept completely replaces the first concept. This problem consists of 120 K data samples and the concept drift occurs after 40K-th samples. This problem is simulated using the periodic hold-out scenario with 1,000 time stamps. Each time stamp involves 1,200 samples where 1,000 samples are used to train the model and the remainder 200 samples serve as the validation samples. The numerical results are presented as the average of 1,000 time stamps.

pENsemble+ delivers encouraging numerical results where it characterizes low sample consumption while producing comparable accuracy. The online active learning scenario is capable of significantly reducing the number of training samples where pENsemble+ only attracts 30% of the total training samples for training process. Moreover, the ensemble merging strategy relieves computational and space complexity by getting rid of redundant classifiers. That is, a classifier sharing strong mutual information with other classifiers can be discarded without substantial loss of generalization power. It is also observed that pENsemble+ with the multivariate Gaussian rule achieves comparable numerical result as that the axis-parallel Gaussian rule. The advantage is seen in terms of the fuzzy rule where the multivariate Gaussian function leads to a more compact and parsimonious rule base than the axis-parallel rule. Nonetheless, the use of multivariate Gaussian function causes slightly higher network parameters to be stored in the memory.

B)*SEA problem*: The SEA problem was developed by Street and Kim [39] and is a popular benchmark problem for concept drift. This problem presents a sudden drift and consists of two input attributes $x_1, x_2$. If the sum of the two attributes fall below the threshold, a sample is classified to class 2, whereas a class 1 is assigned to those higher than the threshold. The concept drift is induced by dramatically changing the thresholds three times during the experiment ( $4 \to 7 \to 4 \to 7$ ) and data samples are uniformly drawn from the range of [0,10]. In our experiment, we use the imbalanced version of this problem introduced by Ditzler and Polikar [17] where the minority class is around 25% of the total data samples. The experiment was carried out in the chunk by chunk mode where each chunk comprises 1000 data samples. The number of chunk is 200 and the drift is applied in every 50 chunk.

SEA problem illustrates that although pENsemble+'s accuracy is comparable to that of pENsemble, it attains much lower sample consumption and comparable structural complexity. Moreover, pENsemble+ is unlike pENsemble which fully operates in the fully supervised manner which demands high operator labelling efforts. pENsemble+ produces similar performance with two different base-classifiers. The learning performance of pENsemble+ including the evolution

[1)] https://www.researchgate.net/profile/Mahardhika_Pratama

of ensemble structure, fuzzy rule, training samples and input attributes are depicted in the supplemental document.

Table 2. Numerical Results of Consolidated Algorithms

| Numerical Example | Evaluation Criteria | pENsemble | Learn++.NSE | Learn++.cde | pClass | pENsemble+ (axis-parallel) | pENsemble+ (multivariate) |
|---|---|---|---|---|---|---|---|
| SEA | Classification Rate | **0.97±0.02** | 0.93±0.02 | 0.93±0.02 | 0.89±0.1 | **0.97±0.02** | 0.97±0.04 |
| | Fuzzy Rule | 4.1±1.8 | N/A | N/A | 6.6±4.2 | 4.2±1.5 | **2.1±0.41** |
| | Input Attribute | **2** | 3 | 3 | 3 | **2** | **2** |
| | Network Parameters | 40.6±18.3 | N/A | N/A | 157.3±101.9 | 41.6±14.95 | **25.1±4.99** |
| | Execution Time | 1.14±0.2 | 1804.2 | 2261.1 | 0.42±0.3 | 0.28±0.04 | **0.27±0.06** |
| | Training Samples | 500 | 500 | 500 | 500 | 106.86±19.46 | **98.7±26.55** |
| | Ensemble Size | 2.03±0.9 | 200 | 200 | N/A | 2.13±0.72 | **1.04±0.21** |
| Hyperplane | Classification Rate | 0.92±0.02 | 0.91±0.02 | 0.9 | 0.91±0.02 | **0.94±0.02** | **0.94±0.02** |
| | Fuzzy Rule | 3.74±0.7 | N/A | N/A | 3.8±1.7 | 2 | **1** |
| | Input Attribute | **2** | 4 | 4 | 4 | **2** | **2** |
| | Network Parameters | 44.8±8.2 | N/A | N/A | 114.9±52.6 | **12** | **12** |
| | Training Samples | 1000 | 1000 | 1000 | 1000 | **241.9±129.6** | 248.9±125.7 |
| | Execution Time | 1.5±0.45 | 926.04 | 2125.5 | 2.7±1.4 | **0.22±0.09** | 0.24±0.1 |
| | Ensemble Size | 1.87±0.34 | 100 | 100 | N/A | **1** | **1** |
| SUSY | Classification Rate | **0.77±0.04** | | | 0.73±0.06 | 0.76±0.04 | **0.77±0.04** |
| | Fuzzy Rule | 3.8±1.5 | | | 1.96±0.26 | 11.2±3.4 | **2** |
| | Input Attribute | **1** | | | 18 | **1** | **1** |
| | Network Parameters | 22.9±9.1 | Terminated | | 748±33.3 | 67.8±19.35 | **12** |
| | Training Samples | 400 | | | 400 | **235.8±13.9** | 319.7±6.02 |
| | Execution Time | **0.29±0.08** | | | 0.79±0.3 | 0.31±0.08 | 0.49±0.25 |
| | Ensemble Size | 1.9±0.7 | | | N/A | 4.3±1.3 | **1** |

C)*Susy problem*: Susy problem is a popular big dataset. It presents a binary classification problem which aims to classify a signal process that produces supersymmetric particles [40] and a background process which does not. This problem has 18 input features in which the first 8 input features are the kinematic properties and the last 10 features are simply the function of the first 10 attributes. It consists of 5-millions data samples where 4.5-millions samples are reserved for the training samples, while the last 500 K samples are used for the testing samples. To simulate data stream environments, data come in batches with 10000 timestamps.

This case study demonstrates the scalability of pENsemble+ for large-scale applications. It exhibits significant improvement over its predecessor, pENsemble, in terms of runtime. It is attributed by the online active learning scenario which brings down the sample consumption to a low level. Local experts can be added and removed on demand and on the fly. Although pENsemble+ works on the chunk by chunk basis, it does not revisit previously acquired data chunk. The memory demand hence remains independent from the total number of data chunks. Furthermore, the ensemble merging scenario reduces the memory complexity and from our numerical examples one can perceive that pENsemble+ has parsimonious and compact network structure. The use of multivariate Gaussian function brings positive effect to alleviate computational and structural burdens of pENsemble+. This result, however, comes at the cost of a slight deterioration of the predictive performance.

Table 3 Experimental Procedure

| *Data stream* | *IA* | *C* | *DP* | *TS* | *TRS* | *TES* | *IF* | *Scenario* |
|---|---|---|---|---|---|---|---|---|
| *SEA* | *3* | *2* | *100000* | *200* | *250* | *250* | *0.13* | *Holdout* |
| *Hyperplane* | *4* | *2* | *120 K* | *100* | *1000* | *250* | *0.012* | *Holdout* |
| *Susy* | *18* | *2* | *5 M* | *10000* | *400* | *100* | *0.085* | *Holdout* |
| *TCM(binary)* | *12* | *2* | *69* | *3* | *10* | *19* | *0.3623* | *Holdout* |
| *TCM(multi-class)* | *12* | *4* | *119* | *5* | *10* | *13* | *0.4202* | *Holdout* |

IA: Input Attributes, C: Classes, DP: Data Points, TS: Time Stamps, TRS: Training Samples, TES: Testing Samples, IF: Imbalance Factor

## VII. CONCLUSION

This paper proposes the novel tool condition monitoring methodology taking advantage of an evolving ensemble fuzzy classifier, pENsemble+. pENsemble+ offers an extension of pENsemble by integrating online active learning scenario and ensemble merging scenario. These two learning components improve the viability of pENsemble for real-world deployment because it can reduce sample consumption, labelling effort and ensemble complexity to modest level. Real-world experiments were carried out where real-world manufacturing data from metal turning and 3D-printing processes were collected. In addition, numerical examples using well-known data streams are provided. It is shown that pENsemble+ delivers encouraging performance in attaining tradeoff between accuracy and complexity. Future work will be directed to study a stacking ensemble architecture for regression problems.

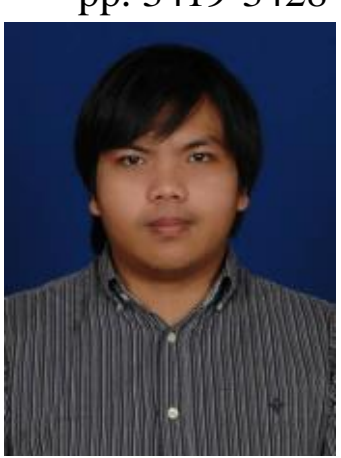

**Mahardhika Pratama** received his PhD degree from the University of New South Wales, Australia in 2014. He is a faculty member at the School of Computer Science and Engineering, Nanyang Technological University, Singapore. He worked as a lecturer at the Department of Computer Science and IT, La Trobe University from 2015 till 2017. Prior to joining La Trobe University, he was with the Centre of Quantum Computation and Intelligent System, University of Technology, Sydney as a postdoctoral research fellow of Australian Research Council Discovery Project. His research interests involve machine learning, computational intelligent, evolutionary computation, fuzzy logic, neural network and evolving adaptive systems.

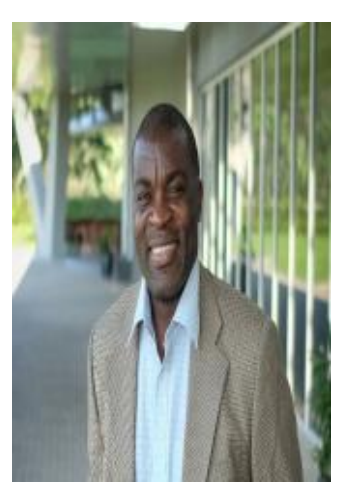

**Eric Dimla** received the MEng (Hons) degree in Mechanical Engineering in 1994 from University College London (University of London) and did a PhD immediately after that on **'Tool condition monitoring using neural networks in metal turning operations'** awarded in June 1998. Assoc. Prof Dimla is Head of School of Science and Technology, RMIT University Vietnam. Prior to joining RMIT Vietnam, he was Dean of the Faculty of Engineering at Universiti Teknologi Brunei (UTB) and Professor of Mechanical Engineering. Preceding joining UTB, he was Academic Leader (Engineering) at Southampton Solent University UK and a Senior Lecturer in CAD and Engineering at Portsmouth University, UK

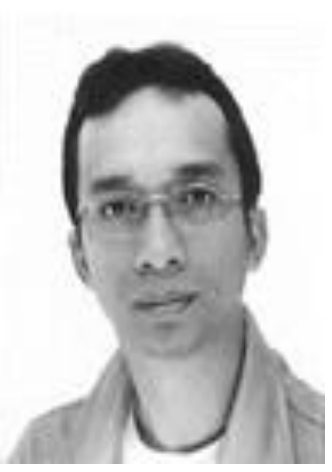

**TEGOEH TJAHJOWIDODO** received the Ph.D. degree from the Katholieke Universiteit Leuven, Belgium, in 2006. He joined the Flanders' Mechatronics Technology Center, Belgium, a research center in the mechatronics area, where he was involved in several projects, mainly in nonlinear dynamics. He is currently an Associate Professor and Assistant Chair of Research at the School of Mechanical & Aerospace Engineering, NTU, Singapore. His research interests are in nonlinear dynamics, modelling, and control.

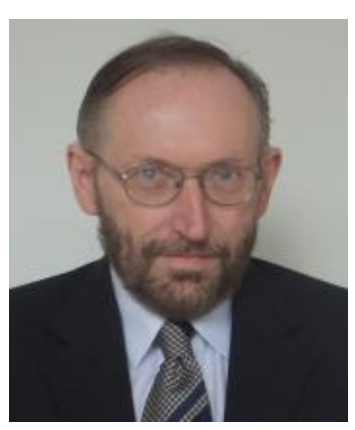

**Witold Pedrycz** (IEEE Fellow, 1998) is Professor and Canada Research Chair (CRC) in Computational Intelligence in the Department of Electrical and Computer Engineering, University of Alberta, Edmonton, Canada. He is also with the Systems Research Institute of the Polish Academy of Sciences, Warsaw, Poland. In 2009 Dr. Pedrycz was elected a foreign member of the Polish Academy of Sciences. In 2012 he was elected a Fellow of the Royal Society of Canada. Witold Pedrycz has been a member of numerous program committees of IEEE conferences in the area of fuzzy sets and neurocomputing. In 2007 he received a prestigious Norbert Wiener award from the IEEE Systems, Man, and Cybernetics Society. He is a recipient of the IEEE Canada Computer Engineering Medal, a Cajastur Prize for Soft Computing from the European Centre for Soft Computing, a Killam Prize, and a Fuzzy Pioneer Award from the IEEE Computational Intelligence Society.

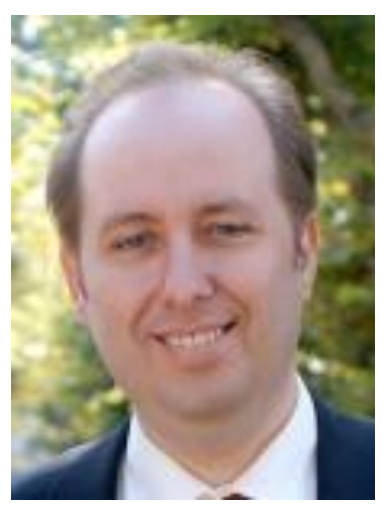

**Edwin Lughofer** received his PhD-degree from the Johannes Kepler University Linz (JKU) in 2005. He is currently Key Researcher with the Fuzzy Logic Laboratorium Linz / Department of Knowledge-Based Mathematical Systems (JKU) in the Softwarepark Hagenberg, see www.flll.jku.at/staff/edwin/. He has participated in several basic and applied research projects on European and national level, with a specific focus on topics of Industry 4.0 and FoF (Factories of the Future). He has published around 200 publications in the fields of evolving fuzzy systems, machine learning and vision, data stream mining, chemometrics, active learning, classification and clustering, fault detection and diagnosis, quality control, predictive maintenance, including 80 journals papers in SCI-expanded impact journals, a monograph on 'Evolving Fuzzy Systems' (Springer) and an edited book on 'Learning in Non-stationary Environments' (Springer).